\documentclass{article}
\usepackage[final,nonatbib]{neurips_2021}
\usepackage[utf8]{inputenc} 
\usepackage[T1]{fontenc}    
\usepackage{hyperref}       
\usepackage{url}            
\usepackage{booktabs}       
\usepackage{amsfonts}       
\usepackage{nicefrac}       
\usepackage{microtype}      
\usepackage{graphicx}
\usepackage{wrapfig}
\usepackage{amsmath}
\usepackage{bbold}
\usepackage{enumitem}

\DeclareMathOperator*{\argmax}{argmax}

\title{On the Value of ML Models}
\author{Fabio Casati and Pierre-Andr\'e No\"el, \\
Element AI, a ServiceNow company
 \And
 Jie Yang\\
 TU Delft}
\date{September 2021}

\begin{document}
\maketitle


\begin{abstract}
We argue that, when establishing and benchmarking Machine Learning (ML) models, the research community should favour evaluation metrics that better capture the value delivered by their model in practical applications. For a specific class of use cases---selective classification---we show that not only can it be simple enough to do, but that it has import consequences and provides insights what to look for in a ``good'' ML model.
\end{abstract}

\section{Introduction}
In business use cases and other practical applications, Machine Learning (ML) models are but components of larger processes, and the value we derive from such models thus depends on the profitability and performances of the workflow as a whole.
However, when determining by which metrics models are to be benchmarked, we believe that the ML research community currently gives very little thoughts as to how value is generated in practice.
While we acknowledge that researchers can't be expected to account for the full diversity and/or complexity of real settings, we believe that value-estimating metrics can be designed to capture high-level commonalities among classes of use cases: benchmarks could (and should!) strive to reflect value.
In this paper, we substantiate these claims for the ubiquitous practical case of selective classification.

\begin{wrapfigure}{r}{0.6\textwidth}
  \centering
  \includegraphics[width=0.6\textwidth]{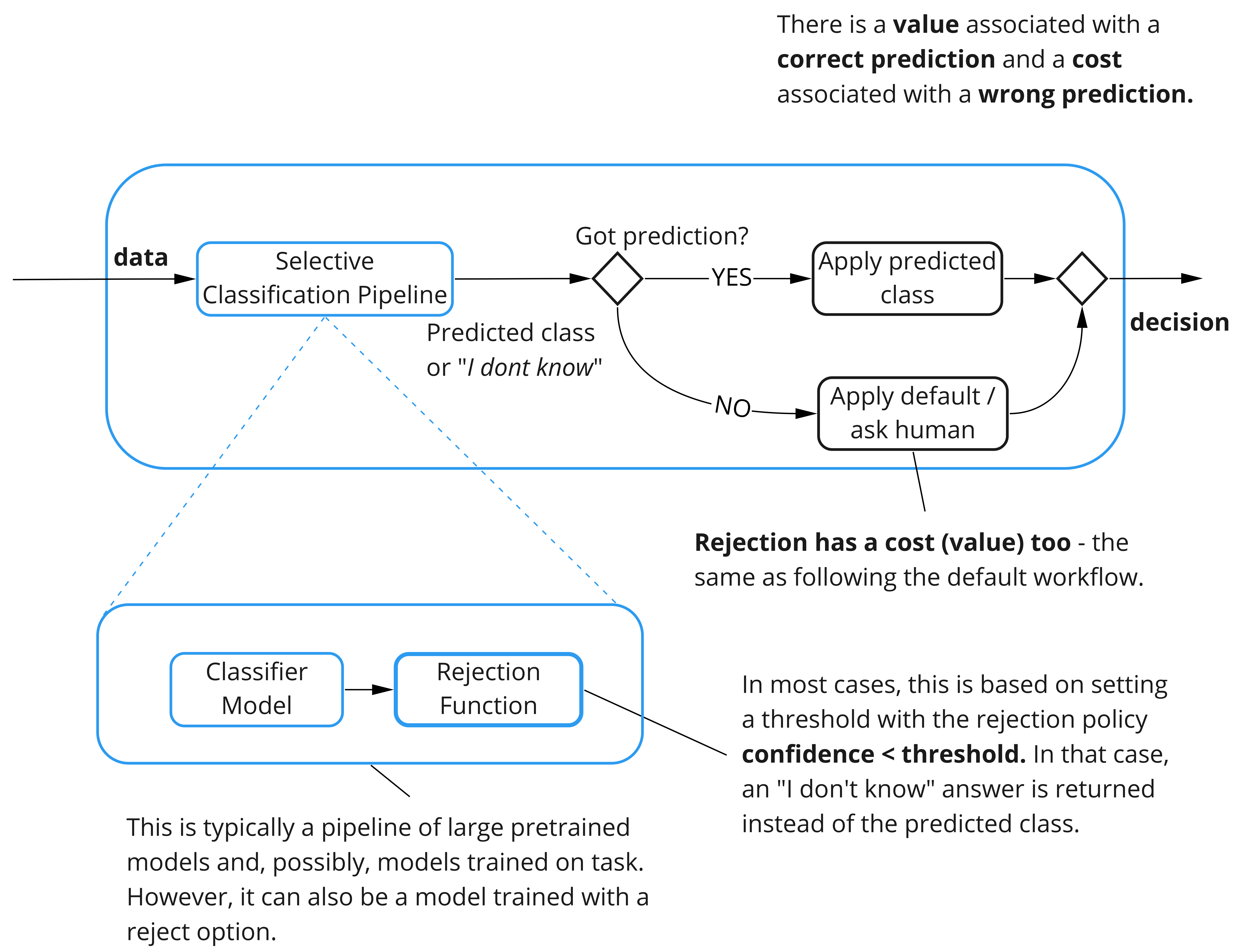}
  \caption{ML Models in Enterprise Workflows. The outcome of the whole workflow---not the prediction of the ML model(s)---is what matters to the users. We advocate for research benchmarks to better account for this reality.}
  \label{fig:wf}
\end{wrapfigure}

Indeed, in the overwhelming majority of enterprise workflows that we are aware of, ML models are deployed with a ``default'' (or ``reject'') option: if the model is ``confident enough'' of a prediction, that prediction is acted on. 
Otherwise a default, safe path is taken (\textit{e.g.}, redirecting chat to a human agent, asking the driver to take the wheel). 
Figure~\ref{fig:wf} illustrates such workflows. 
Here we make these considerations concrete by introducing a special kind of ``selective classifiers'' \cite{bartlett2008classification,fumera2000reject,Selective17,jiang2018trust,pmlr-v119-mozannar20b}, which we call \emph{abstaining classifiers}.
We then posit a simple form for the value derived from an abstaining classifier in an hypothetical workflow. Nontrivial consequences emerge from these basic assumptions, and we argue that these consequences would still play an important role in more realistic contexts: benchmarks not accounting from these phenomenons may paint a picture that has little to do with model value in real use cases.

There has been searches for ``good general purpose metric to use when more specific criteria are not known'' \cite{DataMiningInMetricSpace}, and the literature has many examples where a mismatch between metrics and actual value may bring surprising, important consequences \cite{chybalski2017forecast,pmlr-v80-liu18c,wunderlich2020betting}.
In fact it has long been known \cite{boothe1987comparing} that, given a concrete use case, the overall value of a process should be considered when selecting which model to use (and how to use it). 
Our stance and contributions are complementary: how should researchers benchmark general-purpose models for a broad spectrum of concrete use cases where they may bring value? 
Is such a feat possible at all?

Our results show that there is at least one class of practical concerns---processes with a ``default'' option---that can be distilled into a manageable, actionable metric, with nontrivial consequences backpropagating toward ML research. 
Among these consequences is the observation that ``learning'' does not imply a better accuracy nor a better calibration: given a calibrated model, its value can be increased without altering its prediction (and thus its accuracy) by \emph{learning} a confidence distribution that is more discriminating. 
We hope that future work will show that this is not an isolated case, and that such value-related concerns will play a more important role in the mainstream ML research narrative.

\section{Theory: Estimating the Value of a Model in a Simple Use Case \label{sec:theory}}

This section formalizes how we estimate the value of a classifier that is part of a process with a ``default'' option. We posit rather simple assumptions, and investigate some of their consequences. Questions as to whether those assumptions hold in real use cases are briefly treated in Appendix~\ref{appendix:caveat}.

\textbf{Abstaining classifiers.}
Traditionally, a classifier $\mathbf{f}: X \to [0,1]^C$ uses the input data $\mathbf{x} \in X$ to assign a \emph{confidence} $f_i(\mathbf{x})$ to each class $i \in \{1, \cdots, C\}$, with the understanding that $\argmax_i f_i(\mathbf{x})$ is the \emph{predicted class} with associated confidence $\max_i f_{i}(\mathbf{x})$. We define an abstaining classifier as a function $g: X \mapsto \{0, 1, \cdots, C\}$ mapping the input data $\mathbf{x} \in X$ to a single class $g(\mathbf{x}) \in \{0, 1, \cdots, C\}$, where the special class $0$ indicates that the classifier \emph{abstains} from making a call\footnote{$g$ does not provide \emph{any} ``score'': confidence may be used internally, but it is \emph{not} exposed through $g$'s ``API''.}. Many strategies, including end-to-end training, can be used to obtain an abstaining classifier, but by far the most common approach in applications is to apply a \emph{threshold} $t \in \mathbb{R}$ to a provided traditional classifier $\mathbf{f}$ so that the model returns the traditional predicted class if its confidence exceeds the threshold, and abstains otherwise
\begin{equation}
    g_{\mathbf{f}, t}(\mathbf{x}) = \begin{cases}
        \displaystyle\argmax_{i \in \{1, \cdots, C\}} f_i(\mathbf{x}) & \text{if $\displaystyle \max_{i \in \{1, \cdots, C\}} f_i(\mathbf{x}) \ge t$,}\\
        \displaystyle 0 & \text{otherwise.} \label{eq:f2g}
    \end{cases}
\end{equation}

\textbf{Value.}
There is a benefit (resp. cost) for a correct (resp. wrong) prediction, as well as a cost for following the default path, and these quantities depend on the use case.
We make this concrete for a specific use case $\mathfrak{U}$ by prescribing the value---measured in dollars or other forms of utility---that a customer gets out of an abstaining classifier's prediction. 
Specifically, $\mathfrak{U} = (V_\text{correct}, V_\text{abstain}, V_\text{wrong})$ is a triple whose entries respectively characterize the value of a correct prediction, of an abstention, and of an incorrect prediction.
Furthermore, we assume that these values combine linearly so that, given a dataset $\mathfrak{D} = \left\{(\mathbf{x}_j, y_j) \| j \in \{1, \cdots, N\} \right\}$ composed of $N$ pairs of inputs $\mathbf{x}_j$ and labels $y \in \{1, \cdots, C\}$, we may count the numbers $N_\text{correct}$ of correct predictions, $N_\text{abstain}$ of abstentions, and $N_\text{wrong}$ of incorrect predictions of any abstaining classifier $g$, from which we can establish the total value brought by $v$ in the use case $\mathfrak{U}$ for the dataset $\mathfrak{D}$
\begin{equation}
    V(g, \mathfrak{U}, \mathfrak{D}) = V_\text{correct} \cdot N_\text{correct} + V_\text{abstain} \cdot N_\text{abstain} + V_\text{wrong} \cdot N_\text{wrong} . \label{eq:value}
\end{equation}
For the sake of our argument, we posit that Eq.~\ref{eq:value} is \emph{exactly true}. In fact, we further posit that it is the upper bound to the value that \emph{any model} making the same predictions as $g$---although perhaps returning extra information such as confidences---could possibly achieve in this use case. 
Of course, this is probably never ``exactly true'', and Appendix~\ref{appendix:caveat} discusses some reasons why.
Nonetheless, we believe that our simple assumptions capture important properties of the class of use cases with a ``default'' option, and that their consequences should be treated seriously.

\textbf{Dimensionless formulation.} In the interesting\footnote{Indeed, we would never consider abstention if $V_\text{abstain} < V_\text{wrong}$, and we would always abstain if $V_\text{wrong} < V_\text{abstain} > V_\text{correct}$.} case where $V_\text{wrong} \le V_\text{abstain} < V_\text{correct}$, applying to Eq.~\eqref{eq:value} the change of variables
\begin{equation}
    \mathcal{V}(g, \omega, \mathfrak{D}) = \frac{V(g, \mathfrak{U}, \mathfrak{D}) - \lvert\mathfrak{D}\rvert V_\text{abstain}}{\lvert\mathfrak{D}\rvert (V_\text{correct}-V_\text{abstain})}
    \text{\quad where \quad}
    -\omega = \frac{V_\text{wrong} - V_\text{abstain}}{V_\text{correct}-V_\text{abstain}}
\end{equation}
gives the dimensionless average value per sample
\begin{equation}
    \mathcal{V}(g, \omega, \mathfrak{D}) = \frac{N_\text{correct} - \omega N_\text{wrong}}{N_\text{correct} + N_\text{abstain} + N_\text{wrong}} . \label{eq:dimensionless}
\end{equation}
We can understand Eq.~\eqref{eq:dimensionless} as an empirical average where each correct prediction grants a value of $1$, an abstention grants no value, and a wrong prediction incurs a cost (negative value) determined by the \emph{penalty}\footnote{You may use "$\omega$rong" as a mnemonic device.} parameter $\omega \ge 0$, which \textbf{fully captures the relevant information} from the use case $\mathfrak{U}$.

\textbf{$\omega$-aware \textit{vs} calibrated.} We say of an abstaining classifier that it is $\omega$-aware if it can depend---explicitly, through training or otherwise---on the penalty parameter $\omega$. For example, given a standard classifier $\mathbf{f}$ and defining\footnote{$t_\omega$ is an hyperparameter, hence why we optimize using a different (\textit{e.g.}, validation) dataset $\mathfrak{D}'$.}
\begin{subequations}\begin{equation}
    t_\omega = \argmax_{t \in \mathbb{R}} \mathcal{V}(g_{\mathbf{f}, t}, \omega, \mathfrak{D}') , \label{eq:threshold:uncal}
\end{equation}
then $g_{\mathbf{f}, t_\omega}$ is penalty-aware. In fact, according to our last assumption in introduction, it is the best value you can get out of $\mathbf{f}$. Interestingly, the case $\omega=0$ gives $t_0=0$: in that limit, the model never abstains and $\mathcal{V}(g_{\mathbf{f}, 0}, 0, \mathfrak{D})$ corresponds to the traditional classification accuracy.

Conversely, while the threshold in Eq.~\eqref{eq:threshold:uncal} depends on the classifier $\mathbf{f}$, $\omega$ fully specifies the threshold to be used for all \emph{calibrated} classifiers, \textit{i.e.}, whose confidence score matches the probability for the associated prediction to be correct. Indeed, if the standard classifier $\mathbf{f}^\text{cal.}$ is calibrated with respect to the dataset $\mathfrak{D'}$, then we have
\begin{equation}
    \mathcal{V}(g_{\mathbf{f}^\text{cal}, t}, \omega, \mathfrak{D}') = \int_t^1 \bigl[c - \omega(1-c)\bigr] \rho(c) dc
    \,\text{, which is maximized at }
    t_\omega^\text{cal} = \frac{\omega}{\omega+1} .
    \label{eq:threshold:cal}
\end{equation}\label{eq:threshold}\end{subequations}
Here $\rho: [0,1] \to \mathbb{R}_{\ge0}$ is the Probability Density Function (PDF) for the confidence score given by $\mathbf{f}^\text{cal}$ for an input randomly sampled from $\mathfrak{D}'$. Since $\mathbf{f}^\text{cal}$ is calibrated, $c\,\rho(c)$ is the PDF for its correct classifications at confidence $c$.

If $\mathbf{f}^\text{cal}$ has been obtained by ``calibrating'' $\mathbf{f}$ so that, for any sample, their predicted classes match and the former's confidence is provided by applying a monotonously increasing function $\widetilde{c}: [0, 1] \to [0, 1]$ to the latter's, then Eqs.~\eqref{eq:threshold:uncal} and \eqref{eq:threshold:cal} can be related with $t_\omega^\text{cal} = \widetilde{c}(t_\omega)$. More importantly, both approaches result in the same dimensionless value: \textbf{calibration (limited to a rescaling $\widetilde{c}$) does not improve an abstaining classifier} $g_{\mathbf{f}, t_\omega}$ whose threshold is fixed by Eq.~\eqref{eq:threshold:uncal}.

\textbf{VOC curves.} In analogy with the Receiver Operating Characteristic (ROC) curve often used to characterize binary classifiers, we now introduce the \emph{Value Operating Characteristic} (VOC) curve to characterize abstaining classifiers and, using Eq.~\eqref{eq:f2g}, standard classifiers. The VOC curve of a abstaining classifier $g$ is simply the plot of the dimensionless value per sample $\mathcal{V}(g, \omega, \mathfrak{D}_\text{test})$ as a function of the penalty $\omega$ for a given test dataset $\mathfrak{D}_\text{test}$.

Unless a model $g$ doesn't make a single wrong classification on the provided dataset $\mathfrak{D}_\text{test}$, there exist values of $\omega$ for which $\mathcal{V}(g, \omega, \mathfrak{D}_\text{test})$ is zero or negative. When this happen, $g$ is \emph{useless} in the sense that we may as well enforce the policy ``always abstain'' to achieve a dimensionless value of zero. The quantity $\omega_\text{sup} = \sup \bigl\{\omega \in [0,\infty] \bigm\vert \mathcal{V}(g, \omega, \mathfrak{D}_\text{test}) > 0 \bigr\}$ is the upper bound of the regime where that model has any use for that dataset.

But perhaps the most important property of VOC curves is that, in the limit of an infinite dataset $\mathfrak{D}_\text{test}$, if an abstaining model's VOC curve is everywhere above (or tying with) another model's VOC curve in the interval $0 \le \omega \le \omega_\text{sup}$, then the former model is guaranteed to provide more (or as much) value as the latter for \emph{any} use case: \textbf{it is just ``better''}. It is because of this universality over use cases that we advocate for VOC curves to become an important benchmarking component for all classification problems.

\textbf{Improving standard classifiers.} VOC curves compare models in terms of their value: here we consider how we may alter a standard classifier to ``improve'' its VOC curve. Since calibration cannot affect the VOC curve, we restrict our study to calibrated classifiers without loss of generality. Inspecting the left part of Eq.~\eqref{eq:threshold:cal}, we notice that $\rho$ fully specifies $\mathcal{V}(g_{\mathbf{f}^\text{cal}, \omega(\omega+1)^{-1}}, \omega, \mathfrak{D}')$. We thus consider perturbations of $\rho$ and investigate how they affect the VOC.

If some mass is taken from $\rho$ near some $c_0$ to be ``pushed up'' to higher confidence, this will always result in a better classifier (\textit{i.e.}, improves the value for some ranges of $\omega$, and doesn't make it worse for any others). Conversely, pushing mass down makes it worse (\textit{i.e.}, degrades value for some $\omega$, and cannot make it better for any $\omega$). However, since we consider a calibrated model, this is trivially related to a change in model accuracy $\int_0^1 c\,\rho(c)dc$. What is slightly less trivial is that we can push a fraction greater or equal to $1-c_0$ of the mass around $c_0$ by pushing up a fraction greater or equal to $c_0$, with equality corresponding to the case where accuracy is kept constant.

To make the arguments from the previous paragraph crisper, we introduce the \emph{discrimination} metric $\int_0^1 (\frac{1}{2}-c)^2\rho(c)dc$: it is independent from both accuracy\footnote{To understand why, consider the new distribution $\rho'(c) = \rho(1-c)$.} and calibration\footnote{From our assumption that $\mathbf{f}^\text{cal}$ is already calibrated.} concerns.
It can be understood as the average of two quantities: how far the confidence of good predictions is from $1$, \textit{i.e.}, $\int_0^1 c^2\,\rho(c)dc$, and how far the confidence of bad predictions is from $0$, \textit{i.e.}, $\int_0^1 (1-c)^2\rho(c)dc$. 
\textit{Ceteris paribus}, a classifier with a higher discrimination is preferable to one with a lower one. Higher discrimination does not guarantee a better classifier, but neither does a better accuracy.

\textbf{Learning beyond accuracy and calibration.}
Finally, we introduce, as a thought experiment and/or as an actual ML model, a family of \emph{discriminators} $h: (X \times \{1, \cdots, C\} \times [0,1]) \to [0,1]$ so that if $\hat{\imath}_{\mathbf{f}, \mathbf{x}} = \argmax_{i \in \{1, \cdots, C\}} f_i(\mathbf{x})$ is the top class predicted with confidence $\hat{c}_{\mathbf{f}, \mathbf{x}} = \max_{i \in \{1, \cdots, C\}} f_i(\mathbf{x})$ by $\mathbf{f}$ for the input $\mathbf{x}$, then $h(\mathbf{x}, \hat{\imath}_{\mathbf{f}, \mathbf{x}}, \hat{c}_{\mathbf{f}, \mathbf{x}})$ is the \emph{revised confidence} of the discriminator. With a minor abuse of notation, we can combine the two models to obtain the abstaining classifier
\begin{equation}
    g_{h \circ \mathbf{f}, t}(\mathbf{x}) = \begin{cases}
        \hat{\imath}_{\mathbf{f}, \mathbf{x}} & \text{if $\displaystyle h\left(\mathbf{x}, \hat{\imath}_{\mathbf{f}, \mathbf{x}}, \hat{c}_{\mathbf{f}, \mathbf{x}}\right) \ge t$,}\\
        \displaystyle 0 & \text{otherwise.} \label{eq:discriminator}
    \end{cases}
\end{equation}
By design, $h$ can only affect whether the model abstains or not, and thus cannot affect the accuracy (\textit{i.e.}, the total proportion of correct predictions in the limit $t = \omega = 0$). In fact, when paired with an already calibrated standard classifier $\mathbf{f}^\text{cal}$, any improvement in the VOC of $g_{h \circ \mathbf{f}^\text{cal}, \omega(\omega+1)^{-1}}$ over the one of $g_{\mathbf{f}^\text{cal}, \omega(\omega+1)^{-1}}$ cannot be ``blamed'' on accuracy nor calibration, but should instead relate to discrimination.

Some readers may notice that, in Eq.~\eqref{eq:discriminator}, the discriminator $h$ basically acts as a binary classifier assessing whether $\mathbf{f}$ correctly classifies $\mathbf{x}$ or not. From this perspective, using the traditional binary classifier lingo, $N_\text{correct}$ amounts to the number of ``true positives'', $N_\text{wrong}$ amounts to the number of ``false positives'', and $N_\text{abstain}$ amounts to the conflated sum of the number of ``true negative'' and ``false negative''. The fact that these last two quantities are conflated is a feature of our approach, not a bug: when a model appears in a workflow as a selective classifier, it doesn't matter if the model ``would have been right/wrong'' had it not abstained.

\section{Implications and Considerations}

The assumptions we posit in Sec.~\ref{sec:theory} may apply to a different extent in different use cases: Appendix~\ref{appendix:caveat} lists a few caveats to consider. 
However, notwithstanding those caveats, we believe that it is important for the ML community, both research and enterprise, to consider the notion of ``value''. For the sake of simplicity, and to better ground our discussion, the rest of this section continues with those same assumptions. However, the reader is encouraged to think of broader implications.

\textbf{Limits of existing correctness and calibration metrics.}
Accuracy is most useful as a metric when there are no incentives to abstain, because the cost of being wrong is the same for abstaining (\textit{i.e.}, $\omega$ is zero or close to zero).
When abstaining is a valid option (\textit{i.e.}, $\omega > 0$), a classifier may have greater accuracy (\textit{e.g.}, $0.99$) than a second classifier (\textit{e.g.}, $0.01$), yet the latter may have a higher value than the former.
Similar comments could be formulated for other ``correctness'' metrics (\textit{e.g.}, precision and recall).

Expected Calibration Error (ECE) and similar measures of calibration~\cite{googlecal21} are also limited: they may not provide much information as to the model's discriminating power (\textit{e.g.}, a model whose confidence is a constant corresponding to its accuracy has zero ECE, yet those confidences are not very useful), and improving the ECE by itself may not lead to a more valuable model.
In fact, if we are already fixing $t_\omega$ with Eq.~\eqref{eq:threshold:uncal}, there is no way to achieve additional value through any form of calibration performing a monotonically increasing rescaling of the confidence.

Interestingly, having access to a model, regardless of its ``correctness'' and calibration, is almost always better than not having a model at all. Indeed, as long as ranking the predictions in terms of confidence would end up with a sufficient density of correct predictions on the top, then we may use Eq.~\eqref{eq:threshold:uncal} to fix a non-trivial threshold $t_\omega < 1$.

\textbf{The importance of confidence and confidence distributions.}
We argue that the confidence measures of classifiers play a key role in the value of a model (\textit{i.e.}, they are not ``just'' a way to select the predicted class), and that studying (empirical) confidence distributions---the probability distribution or proportion of items with a given confidence---reveals useful information on the quality of a model. 

\begin{wrapfigure}{l}{0.5\textwidth}
  \includegraphics[width=0.5\textwidth]{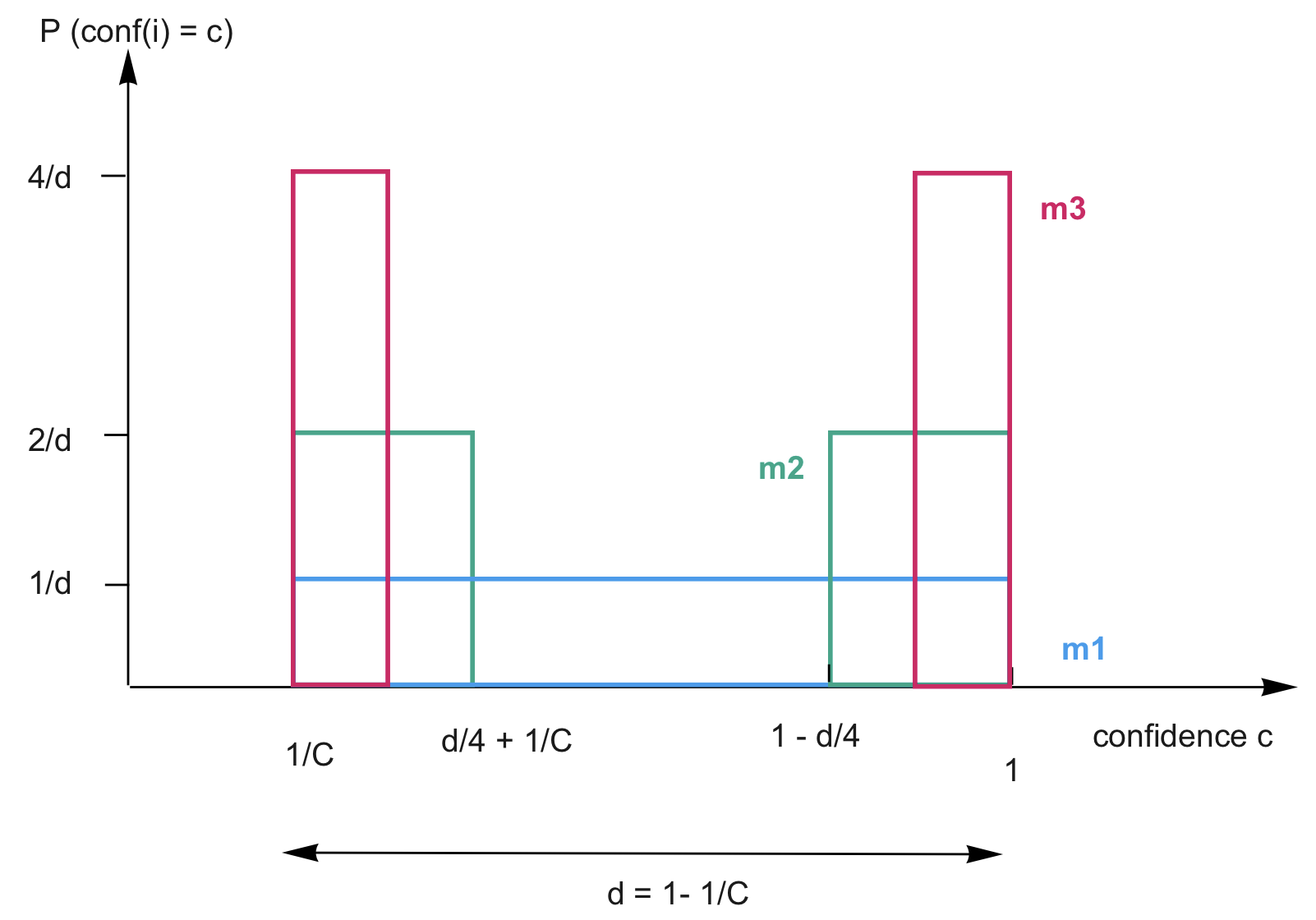}
  \caption{Confidence probability distributions}
  \label{fig:cdistr}
\end{wrapfigure}

To see this, consider the three confidence distributions $\rho$ in Fig~\ref{fig:cdistr}, representing a model that predicts one among C classes (so that the minimal confidence for each prediction is 1/C). 
Assuming that they are perfectly calibrated, the three models have the same accuracy and, by hypothesis, the same calibration. However, the red classifier model $m3$ will always have equal of better value, no matter the cost. 

What makes $m3$ more valuable is that it pushes confidence mass towards the edges, which is both good (many calibrated high confidence predictions, on which the model is truly pretty sure about) and bad (many calibrated low confidence predictions, on which the model is truly unsure about), so that the overall distribution of confidence results in a better model.
This type of ``shove away from the center'' is discussed at the end of Sec.~\ref{sec:theory}. One way to summarize those discussions is that if $m1$, $m2$ and $m3$ were snapshots taken at increasing epochs in a training algorithm, we could safely say that the model is \textbf{learning}, and that what it learns is \textbf{valuable}.
Confidence is not only a mean to an end, it is an end of its own.

To some extent research on unknown unknowns focuses on this~\cite{liu2020towards}.
However, \textbf{(i)} the problem of \textit{unknown knowns} may, depending on $\omega$, be as important, and \textbf{(ii)} the typical definition of unknown unknown (``high confidence errors'')  depends on knowing what ``high confidence'' means, which depends on the threshold and cost. 
This means that we are back to value functions even when we want to explore the unknown unknowns and unknown knowns space.

\textbf{Active Learning and Uncertainty Sampling.}
A question that comes from the above discussion is whether \textit{uncertainty sampling} \cite{aggarwal2014active} is a proper active learning strategy in all cases, or if there are cases where the opposite---``certainty sampling''---would be more valuable. Indeed, for calibrated distributions, we may be more interested to know what is happening above (or not too far below) the threshold $\omega(\omega+1)^{-1}$.

A natural quantity of interest is the \emph{VOC's Area Under the Curve} (VOC AUC) $\int_0^\infty \max\bigl(\mathcal{V}(g, \omega, \mathfrak{D}), 0\bigr) d\omega$. \textit{Ceteris paribus}, we likely prefer a model whose VOC AUC is higher than another one. However, it would probably be more informative to report the VOC AUC for different ranges of values of $\omega$, \textit{e.g.}, $[0,1)$ and $[1,\omega_\infty)$. How many such ranges are desirable and at which values of $\omega$ should we break them are questions that would need to be answered empirically.


\appendix

\section{Some Caveats of our Definition of ``Value''\label{appendix:caveat}}

This Appendix is an non-exhaustive list of possibles ways that the assumptions posited in introductions may not exactly correspond to the reality of the use case.

We posited that the per-sample value was solely dependent on the ``true label'' $y$ and on the prediction of the form $\{0, 1, \cdots, C\}$ (\textit{i.e.}, either a class or $0$ for ``abstain'').
\begin{itemize}
    \item \textbf{Value of confidence.} Value could be derived from the prediction's score in a manner more intricate than whether the prediction should be used or not. For example, a low-ish scored prediction could be used to automatically pre-fill a box in a, but then adding a special flab prompting the user to confirm the value.
    \item \textbf{Class-dependent value.} The value of ``correct'' and ``wrong'' predictions could instead be some $\mathbb{R}^{C \times C}$ matrix whose element $V_{ij}$ is the value of predicting $i$ if the correct answer is $j$ (\textit{i.e.}, the kind of information captured by a confusion matrix). The value of abstaining could similarly depend on the correct label $j$.
    \item \textbf{Ranking/alternatives.} If the default behaviour is that the user must select the correct category in a drop-down menu of 1000 elements, providing a top-10 containing the correct entry at a position different than the first may be more desirable.
    \item \textbf{Nonlinearity.} Equation~\eqref{eq:value} presumes that value scales linearly with count, but nonlinear scaling is also possible. Indeed, different externalities that were not accounted for in the calculation of the cost $V_\text{wrong}$ may begin to matter when proportion of ``wrong'' prediction dominates: loss of customer trust, need to hire new staff, supply-chain failure, \textit{etc.} Such concerns could induce a ``dip'' in the VOC curve at low $\omega$.
\end{itemize}

Please note that our argumentation in the main text is made in view of the caveats listed here: we are aware of these failings, yet we believe our main message to hold true. In particular, focusing too much on the above caveats has caveats of its own.
\begin{itemize}
    \item ``Value'', in a form or another, should play a central role in model benchmarking (and it currently doesn't).
    \item For classification problems, our formulation in terms of $\omega$ captures important concerns that are currently neither properly handled by the research community nor the enterprise.
    \item Classification problems are only one example among many.
\end{itemize}

\end{document}